\documentclass{article}
\usepackage{PRIMEarxiv}
\usepackage[utf8]{inputenc}
\usepackage[T1]{fontenc}
\usepackage{amsmath,amssymb,amsfonts,mathrsfs}
\usepackage{graphicx}
\usepackage{booktabs,multirow,array,tabularx,makecell,threeparttable}
\usepackage[table]{xcolor}
\definecolor{ResultAccent}{HTML}{BF5D55}
\usepackage{caption,subcaption}
\usepackage{float,placeins}
\usepackage{microtype}
\usepackage{cite}
\usepackage[hidelinks]{hyperref}
\usepackage{url}
\usepackage{cleveref}
\hypersetup{
  pdftitle={FastRoverVLA: Fast Vision-Language-Action Execution for Off-Road Autonomous Driving},
  pdfauthor={Changtao Zhao, Yingxue Guan, Li Wang, Shijie Zhao, Yuhang Zhang, Shengkai Zhang, Qing Wang, Ziying Song}
}

\newlength{\ResultMetricWidth}
\newlength{\ResultLabelWidth}
\newlength{\ResultBackboneWidth}
\newlength{\ResultExecutionWidth}
\newlength{\ResultSystemMetricWidth}
\newcolumntype{L}[1]{>{\raggedright\arraybackslash}p{#1}}
\newcolumntype{C}[1]{>{\centering\arraybackslash}p{#1}}
\newcommand{\ResultTableStyle}{%
  \small
  \renewcommand{\arraystretch}{1.00}%
  \setlength{\tabcolsep}{3pt}%
  \setlength{\aboverulesep}{2pt}%
  \setlength{\belowrulesep}{2pt}%
  \setlength{\heavyrulewidth}{0.7pt}%
  \setlength{\lightrulewidth}{0.4pt}%
  \setlength{\cmidrulewidth}{0.35pt}%
  \setlength{\cmidrulekern}{4pt}%
  \setlength{\ResultMetricWidth}{0.075\textwidth}%
  \setlength{\ResultLabelWidth}{\dimexpr\textwidth-8\ResultMetricWidth-16\tabcolsep\relax}%
  \setlength{\ResultBackboneWidth}{0.5\ResultLabelWidth}%
  \addtolength{\ResultBackboneWidth}{-\tabcolsep}%
  \setlength{\ResultExecutionWidth}{\ResultBackboneWidth}%
  \setlength{\ResultSystemMetricWidth}{\dimexpr\textwidth-\ResultLabelWidth-12\tabcolsep\relax}%
  \setlength{\ResultSystemMetricWidth}{\dimexpr\ResultSystemMetricWidth/6\relax}%
  \captionsetup{skip=4pt}%
}

\title{FastRoverVLA: Fast Vision-Language-Action Execution for Off-Road Autonomous Driving}
\author{
  \textbf{Changtao Zhao}$^{1}$, \textbf{Yingxue Guan}$^{1}$, \textbf{Li Wang}$^{1,2,*}$, \textbf{Shijie Zhao}$^{3}$,\\
  \textbf{Yuhang Zhang}$^{3}$, \textbf{Shengkai Zhang}$^{4}$, \textbf{Qing Wang}$^{5}$, \textbf{Ziying Song}$^{5,\dagger}$\\[5pt]
  {\normalfont\small $^{1}$School of Mechanical Engineering, Beijing Institute of Technology, Beijing 100081, China}\\
  {\normalfont\small $^{2}$Chongqing Innovation Center, Beijing Institute of Technology, Chongqing 401120, China}\\
  {\normalfont\small $^{3}$School of Mechanical Engineering and Intelligent Equipment, Chang'an University}\\
  {\normalfont\small $^{4}$School of Electronic and Information Engineering, Beijing Jiaotong University}\\
  {\normalfont\small $^{5}$School of Artificial Intelligence (School of Software), Yanshan University}\\[4pt]
  {\normalfont\small $^{*}$Corresponding author: Li Wang (\texttt{wangli\_bit@bit.edu.cn})}\\
  {\normalfont\small $^{\dagger}$Project lead: Ziying Song}
}

\begin{document}
\raggedbottom
\maketitle

\begin{abstract}
Off-road vision-language-action (VLA) models generate scene-conditioned plans, but frequent trajectory updates can repeatedly invoke expensive multimodal inference. Reusing a prediction reduces this demand while leaving its commands exposed to changes in vehicle motion. FastRoverVLA addresses this mismatch by decoupling visual plan generation from motion-conditioned execution. Its training-free interface maintains decoded plans between visual predictions. A speed--curvature action buffer is aligned with measured motion at each update, integrated into a local trajectory, and advanced by consuming the current action. The corrected remainder is written back to the buffer, preserving motion corrections across successive updates. A bounded refresh rule normalizes speed and curvature variations within a newly decoded plan and assigns a short or long reuse budget according to the resulting plan-roughness score. The interface requires neither backbone features nor an additional learned driving policy. Over a 3~s horizon, independent real-request evaluation achieves average displacement error (ADE) and final displacement error (FDE) of 0.328/0.596~m on ORAD-3D and 0.266/0.526~m on RELLIS-3D, reducing FDE by 18.0\% and 36.7\% at request rates of 6.0\% and 4.8\%, respectively. These runs retain lower errors across four backbones and both datasets and reduce Qwen3-VL request-processing cost by factors of 16.6 and 20.8 under serial replay, while mechanism and motion-only controls identify persistent state correction as the main source of the observed accuracy benefit. The source code and evaluation materials are publicly available at \url{https://github.com/ADEPT-ORG/FastRoverVLA}.

\end{abstract}


\section{Introduction}

Off-road driving requires motion decisions in environments without fixed lane structure, where surface conditions, visibility, and feasible routes can change within a journey~\cite{rellis3d2020,orfd2022,orad3d2025}. Vision-language-action (VLA) models connect visual observations and language-based reasoning to numerical actions or trajectories~\cite{lmdrive2024,drivelm2024,drivevlm2024,emma2024}. Their outputs must also accommodate the vehicle motion that follows prediction. A scene-conditioned plan can remain relevant across several trajectory updates, while its speed and curvature commands require continued adjustment to the measured state.

Coupling every trajectory update to a new prediction repeatedly invokes multimodal inference~\cite{autovla2026,orion2026,drivemoe2026,recogdrive2026}. Efficient encoders and inference pipelines reduce the cost of individual calls~\cite{fastvlm2025,flashdrive2026}, while slow--fast systems combine expensive reasoning with a faster planning path~\cite{drivevlm2024,adadrive2025}. Action-chunk methods address continuity between predicted sequences through temporal aggregation, constrained generation, or learned correction~\cite{act2023,rtc2025,remac2026iclr}. These developments leave a specific execution problem at the decoded-action interface: retaining the motion corrections accumulated after a plan has been generated, without another learned planner or access to the generative sampler.

\begin{figure}[!htbp]
    \centering
    \includegraphics[width=0.72\linewidth,pagebox=cropbox]{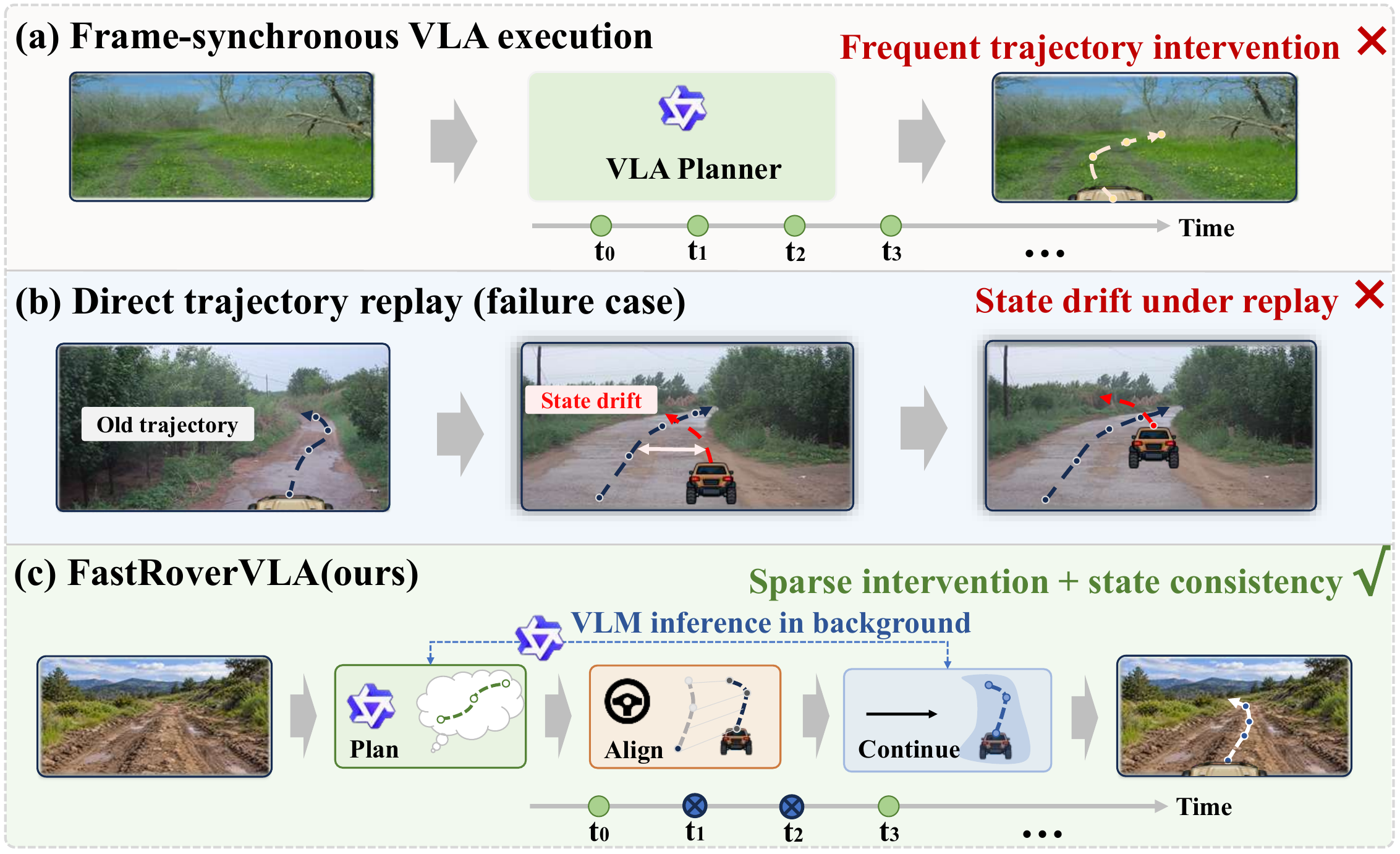}
    \caption{\textbf{Motivation of FastRoverVLA.}
    (a) Frame-synchronous prediction.
    (b) State mismatch during direct reuse.
    (c) Recurrent state alignment and write-back between plan refreshes. The separated paths depict distinct planning and trajectory-update rates; the arrows indicate scheduling dependencies.}
    \label{fig:motivation}
\end{figure}

A retained plan must account for two changes during reuse: commands are consumed as time advances, and measured motion changes the state to which the remaining commands must be matched. Direct replay advances the sequence without resolving this state mismatch. Realigning the original plan at each update incorporates current motion but discards previous corrections. Carrying the corrected remainder into the next update preserves both the unconsumed actions and the motion corrections retained in them (Fig.~\ref{fig:motivation}).

FastRoverVLA decouples visual plan generation from recurrent execution around a frozen vision-language model (VLM) and a deterministic action adapter. The name combines ``Fast'' for reduced VLM inference demand, ``Rover'' for off-road driving, and ``VLA'' for vision-language-action planning. Visual inference refreshes the action buffer, while recurrent execution maintains motion-conditioned trajectories between refreshes. Each update aligns the retained speed--curvature actions with measured ego motion, generates a local trajectory, consumes the current action, and writes the corrected remainder back to the buffer. A bounded rule uses variation within the refreshed plan to assign its reuse budget. The resulting interface separates visual plan generation from motion-conditioned trajectory updates without modifying the backbone or training an auxiliary driving policy.

Evaluation on ORAD-3D~\cite{orad3d2025} and RELLIS-3D~\cite{rellis3d2020} separates three questions: whether decoded plans benefit from persistent correction, whether visual initialization adds information beyond motion history, and whether sparse refresh reduces measured inference demand. Shared-prediction controls isolate the maintenance operations; motion-only and matched-image controls examine initialization. Backbone transfer, disjoint-sequence evaluation, and a past-only heading check test the stability of the execution effect. Independent real-request runs measure its computational consequence under serial replay.

The main contributions are:
\begin{itemize}
    \item FastRoverVLA introduces an action-level execution interface for off-road VLA planning. Maintaining a decoded plan as a recurrent state separates visual prediction from motion adaptation, allowing local trajectory updates between model requests without an additional learned driving policy.
    \item The interface combines state alignment, action consumption, and correction write-back to preserve measured-motion updates in the retained plan. A bounded refresh rule assigns reuse budgets from normalized action variation, controlling inference demand through the same decoded-action representation.
    \item Controlled replay and independent real-request evaluation cover two datasets and four backbones. In real-request evaluation, Qwen3-VL reduces 3~s FDE by 18.0\% and 36.7\% relative to every-update execution, while serial request-processing cost falls by factors of 16.6 and 20.8, respectively.
\end{itemize}

\section{Related Work}

\subsection{Vision-Language-Action Planning}

Robotic VLA models such as RT-2 and OpenVLA~\cite{rt2_2023,openvla2025} connect vision-language representations to actions. Driving planners UniAD and VAD~\cite{uniad2023,vad2023} emphasize planning-oriented scene representations. LMDrive, DriveLM, DriveVLM, and EMMA~\cite{lmdrive2024,drivelm2024,drivevlm2024,emma2024} incorporate language instructions, structured understanding, or hierarchical reasoning. AutoVLA, ORION, ReCogDrive, and SimLingo~\cite{autovla2026,orion2026,recogdrive2026,renz2025simlingo} connect language or reasoning with numerical driving actions; DriveMoE~\cite{drivemoe2026} specializes vision and action experts. ColaVLA, WAM-Flow, and Latent-CoT-Drive~\cite{colavla2026,wamflow2026,latentcot2026} explore latent reasoning, parallel decoding, or world-model-based planning. Complementary work improves the information and constraints used for planning: FocalAD~\cite{sun2026focalad} models decision-critical local interactions, and GraphBEV++~\cite{song2026graphbevplusplus} aligns camera--LiDAR features. GraphWorld and DriveFuture~\cite{song2026graphworld,hong2026drivefuture} use latent world-state or future-aware conditioning. DriveWorld-VLA~\cite{jia2026driveworldvla} integrates world modeling with VLA planning, while MindDrive~\cite{sun2025minddrive} evaluates world-model-generated candidates with a VLM. GuideFlow~\cite{liu2026guideflow} constrains flow matching, and DIVER~\cite{song2026diver} applies reward-guided diffusion to multimodal planning. These advances improve scene-conditioned plan generation, whereas decoded actions must also remain consistent with the changing execution state. FastRoverVLA addresses this execution stage through deterministic measured-motion updates to speed--curvature actions. Its frozen planner supplies new visual plans, while a retained action buffer preserves corrections between predictions, separating plan maintenance from encoder, reasoning, and decoding design.

\subsection{Terrain-Aware Off-Road Planning}

Off-road planning couples geometric clearance with physical traversability. BADGR and STEP~\cite{badgr2020,step2021} learn navigation affordances or combine stochastic traversability with risk-aware planning. How Does It Feel?, V-STRONG, and RoadRunner~\cite{hdif2023,vstrong2024,roadrunner2024} learn terrain costs or traversability from visual, proprioceptive, or LiDAR observations. TRAIL~\cite{trail2026} supports joint path and speed optimization through continuous terrain representations; terrain-adaptive MPC~\cite{yu2025terrainadaptive} incorporates vehicle--terrain interactions and online parameter estimates. OFF-EMMA~\cite{offemma2026} uses visual prompts and self-consistent reasoning for off-road trajectory generation. RELLIS-3D, ORFD, and ORAD-3D~\cite{rellis3d2020,orfd2022,orad3d2025} provide evaluation settings with rugged terrain and varied visibility. These approaches determine where and how to move, but measured vehicle motion can require adjustment without changing the selected route or terrain assessment. FastRoverVLA complements terrain interpretation by updating decoded actions between visual predictions, maintaining motion consistency without requiring new terrain labels or traversability estimates at every update.

\subsection{Efficient and State-Consistent Execution}

Temporal planners exploit cross-frame information: MomAD~\cite{song2025momad} carries trajectory and perception momentum, MomADv2~\cite{song2026momadv2} retains command-consistent memory and uses flow refinement, and MomWorld~\cite{song2026momworld} propagates latent future scene states. Efficiency methods improve visual encoding or model size~\cite{fastvlm2025,vldrive2025}, reduce reasoning~\cite{autovla2026,nord2026,cfvla2026}, or accelerate action generation and inference~\cite{linkvla2026,flashdrive2026}. Slow--fast systems combine infrequent VLM planning with trajectory refinement~\cite{drivevlm2024}, learn adaptive activation and fusion~\cite{adadrive2025}, or schedule query, cache, and drop decisions~\cite{asscg2026}. These mechanisms determine when guidance arrives, while action maintenance determines how decoded sequences evolve during execution. ACT~\cite{act2023} ensembles overlapping predictions, and Diffusion Policy~\cite{diffusionpolicy2023} generates receding-horizon action sequences. RTC~\cite{rtc2025} asynchronously generates chunks while preserving committed actions and constraining overlap through inpainting; REMAC~\cite{remac2026iclr} learns within-chunk corrections and preserves prefixes across chunks. FastRoverVLA instead deterministically aligns a retained speed--curvature sequence with measured motion and writes corrections back for subsequent updates. Its bounded refresh rule admits new visual plans, while intermediate updates require neither sampler access nor an auxiliary learned planner. The focus is persistent state correction after decoding, distinct from learned temporal planning, query scheduling, or new-chunk generation.

\section{Methodology}
\label{sec:method}

\begin{figure}[!htbp]
    \centering
    \includegraphics[width=\textwidth,pagebox=cropbox]{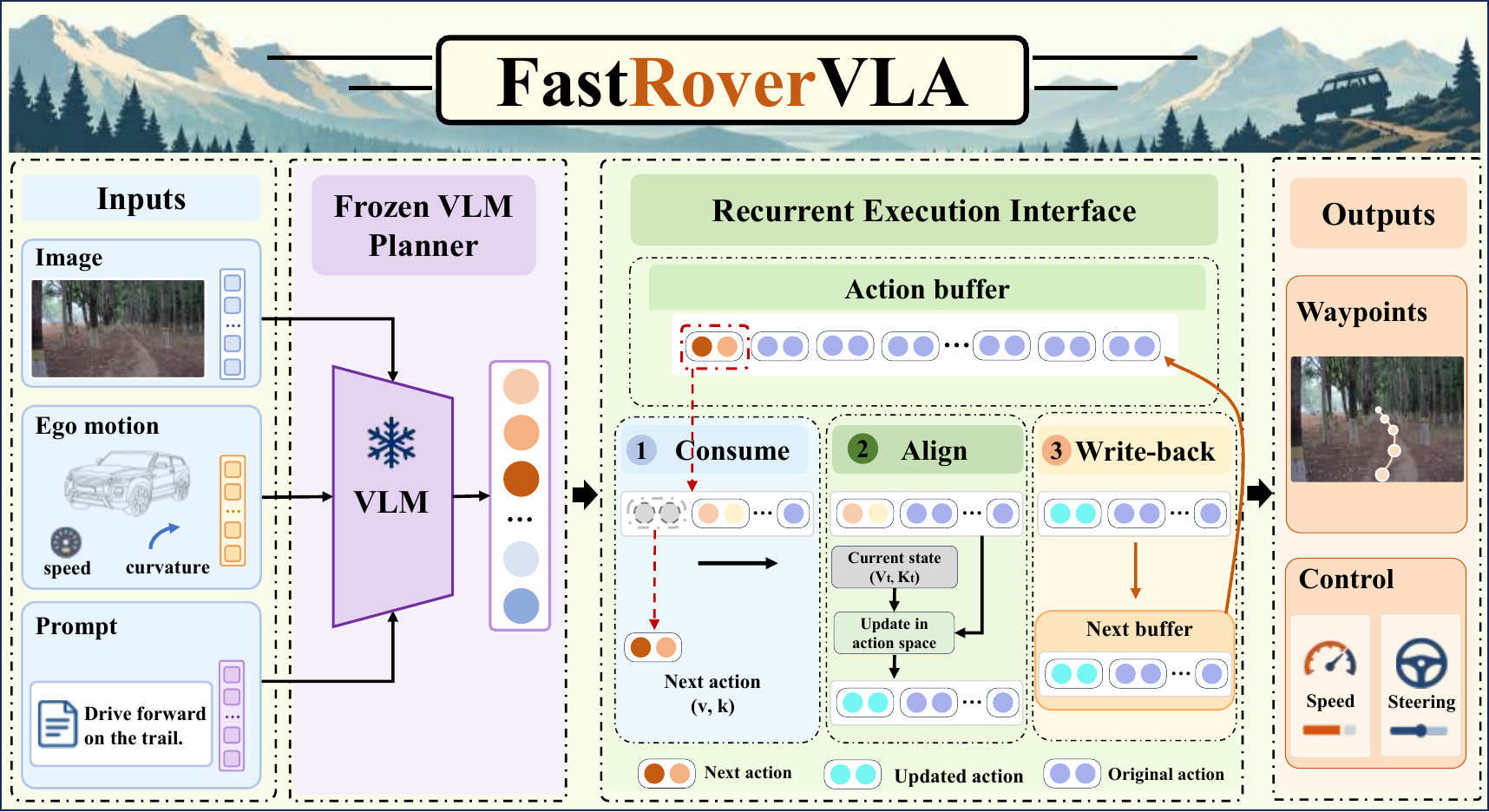}
    \caption{\textbf{FastRoverVLA framework.} A frozen VLM initializes a speed--curvature buffer from images, ego motion, and an instruction. Consume--align--write-back updates produce integrated waypoints between budgeted refreshes (Section~\ref{sec:bounded_refresh}). The control panel denotes a potential downstream interface; evaluation uses waypoints and logged states, without a vehicle controller.}
    \label{fig:method}
\end{figure}

FastRoverVLA maintains a speed--curvature action buffer between visual predictions (Fig.~\ref{fig:method}). A refresh initializes the buffer from the current image, ego motion, and instruction. Subsequent updates align its actions with measured motion and retain the corrected remainder until the reuse budget expires. A plan denotes the action sequence, and a trajectory denotes the positions obtained by integrating that sequence.

\subsection{Action-Buffer Initialization}
\label{sec:buffer_initialization}

Buffer initialization provides a common action representation for fresh predictions and retained plans. Let \(t\) index trajectory updates, \(h\) index future actions, and \(H\) denote the output horizon length. At a refresh, the frozen VLM and its deterministic adapter return \(L_t\leq H\) parsed actions, \(\widetilde A_t=\{(\widetilde v_{t,h},\widetilde\kappa_{t,h})\}_{h=1}^{L_t}\), where \(\widetilde v\) and \(\widetilde\kappa\) denote speed and curvature. A short output is completed by repeating its last action; an empty output uses measured speed and curvature throughout. The completed plan is
\begin{equation}
\hat{A}_t=
\{(\hat{v}_{t,h},\hat{\kappa}_{t,h})\}_{h=1}^{H},
\end{equation}
where \(\hat{v}_{t,h}\) and \(\hat{\kappa}_{t,h}\) are the predicted speed and curvature at horizon step \(h\).

Let \(a_t^0=(v_t,\kappa_t)\) contain measured speed and curvature, and let \(r_t=1\) indicate a VLM refresh (\(r_t=0\) denotes reuse). The active plan is
\begin{equation}
C_t(h)=
\begin{cases}
\alpha \hat{A}_t(h)+(1-\alpha)a_t^0, & r_t=1,\\
\operatorname{Fill}(B_{t-1},H)(h), & r_t=0,
\end{cases}
\end{equation}
where \(B_{t-1}\) stores the corrected remainder from the preceding update and \(\alpha\in[0,1]\) weights the fresh plan. \(\operatorname{Fill}(B,H)\) extends a nonempty buffer to \(H\) actions by repeating its last action. An empty buffer triggers a refresh. Thus, horizon completion uses the available action state without an additional visual prediction.

\subsection{State-Consistent Recurrent Execution}
\label{sec:recurrent_execution}

Changes in vehicle motion create a mismatch between the retained commands and the current state. Each action is therefore aligned with measured speed and curvature:
\begin{equation}
\bar{a}_{t,h}
=
\beta C_t(h)+(1-\beta)a_t^0,
\qquad h=1,\ldots,H,
\end{equation}
where \(\beta\in[0,1]\) weights the active plan and \(\bar{a}_{t,h}=(\bar{v}_{t,h},\bar{\kappa}_{t,h})\) is the aligned action. The fresh-plan weight \(\alpha\) applies only at refresh, whereas \(\beta\) applies at every update, including refresh. Alignment operates in action space, adjusting speed and curvature before integration.

Starting from measured speed, a bound \(\delta_v\geq0\) limits the change between consecutive speed commands. Speeds are clipped to nonnegative values, and curvature is unchanged by this bound. The corrected plan \(A_t^{*}=\{(v_{t,h}^{*},\kappa_{t,h}^{*})\}_{h=1}^{H}\) is integrated into an ego-relative trajectory,
\begin{equation}
Y_t=\Phi(A_t^{*},\Delta t),
\end{equation}
where \(\Phi\) denotes planar integration and \(\Delta t\) is the nominal logged-state and integration interval. Position \((x_h,y_h)\) and heading \(\psi_h\) are measured relative to the vehicle pose at the current update, with \((x_0,y_0,\psi_0)=(0,0,0)\). Omitting index \(t\) for clarity,
\begin{equation}
\begin{aligned}
x_h &= x_{h-1}+v_h^*\cos(\psi_{h-1})\Delta t,\\
y_h &= y_{h-1}+v_h^*\sin(\psi_{h-1})\Delta t,\\
\psi_h &= \psi_{h-1}+\kappa_h^*v_h^*\Delta t.
\end{aligned}
\end{equation}
The output \(Y_t\) contains the resulting \(H\) positions under planar kinematic integration.

After consuming the first action, \(\operatorname{Tail}(\cdot)\) removes it and stores the remaining \(H-1\) corrected actions:
\begin{equation}
B_t=\operatorname{Tail}(A_t^{*}).
\end{equation}
Writing back the corrected remainder makes subsequent alignment depend on the motion corrections retained in the buffer as well as the current measured state. In the cyclic consume--align--write-back view, consumption at the preceding update has already removed the previous interval's action before current alignment begins. During logged replay, subsequent measured states are supplied by the recorded sequence.

\subsection{Bounded VLM Refresh}
\label{sec:bounded_refresh}

\begin{figure}[!htbp]
    \centering
    \includegraphics[width=0.70\linewidth,pagebox=cropbox,keepaspectratio]{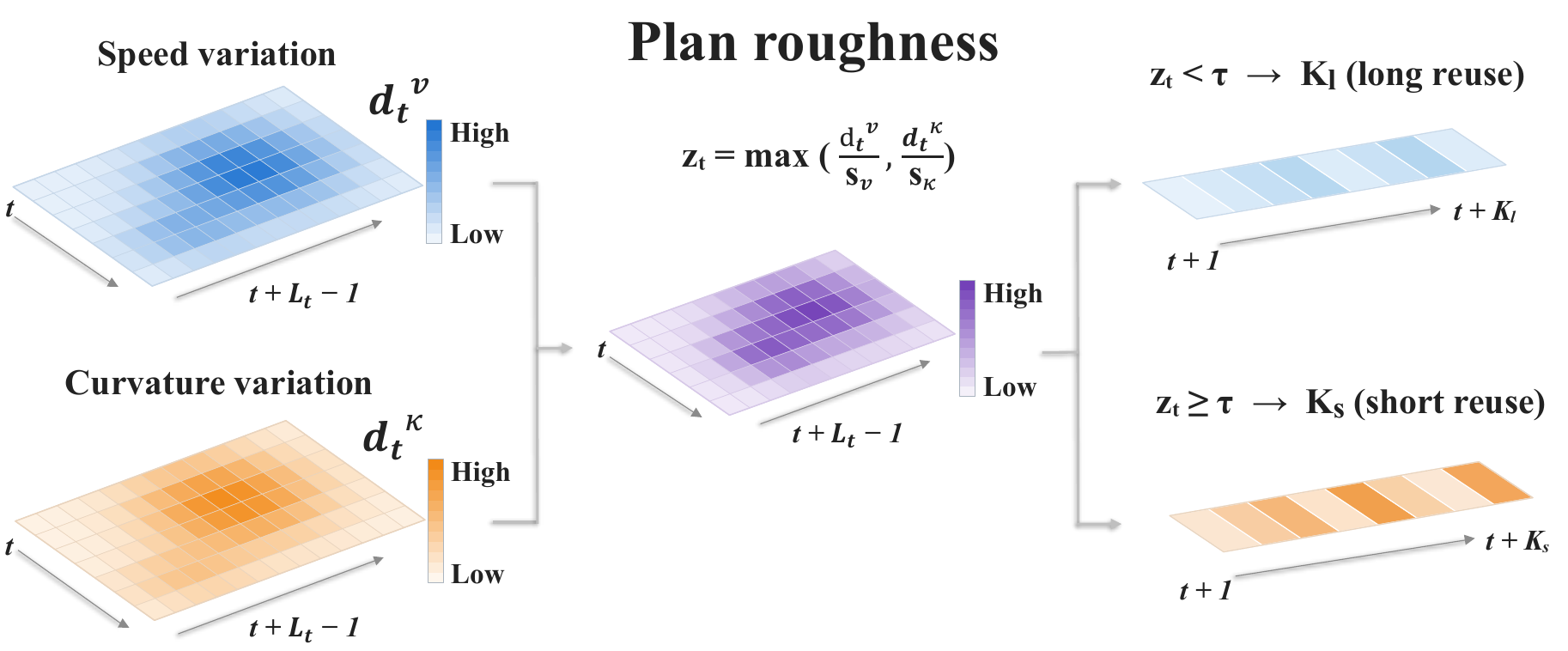}
    \caption{Plan-variation-based bounded refresh. Speed and curvature variations are normalized and aggregated into the plan-roughness score $z_t$. The threshold $\tau$ selects a short reuse budget $K_s$ for higher variation and a long reuse budget $K_l$ otherwise.}
    \label{fig:bounded_refresh}
\end{figure}

The refresh rule controls inference demand through variation in the decoded actions. Higher-variation plans receive shorter reuse budgets, while lower-variation plans can be retained longer. For \(L_t\geq2\), the mean absolute changes in speed and curvature, \(d_t^v\) and \(d_t^\kappa\), are computed before horizon completion:
\begin{equation}
\begin{aligned}
d_t^v &=
\frac{1}{L_t-1}\sum_{h=2}^{L_t}
|\widetilde{v}_{t,h}-\widetilde{v}_{t,h-1}|,\\
d_t^\kappa &=
\frac{1}{L_t-1}\sum_{h=2}^{L_t}
|\widetilde{\kappa}_{t,h}-\widetilde{\kappa}_{t,h-1}|,
\end{aligned}
\end{equation}
The normalized plan-variation score is
\begin{equation}
z_t=
\max
\left(
\frac{d_t^v}{s_v},
\frac{d_t^\kappa}{s_\kappa}
\right),
\end{equation}
where \(s_v\) and \(s_\kappa\) are the validation 90th percentiles of the corresponding raw features, lower-bounded by \(10^{-4}\). Before normalization, both features are set to zero for a single parsed action and to one for an empty parsed list. The score \(z_t\), termed plan roughness, summarizes action variation rather than terrain conditions. The surfaces in Fig.~\ref{fig:bounded_refresh} are schematic representations of these scalar quantities.

The reuse budget is
\begin{equation}
K_t=
\begin{cases}
K_s, & z_t\geq\tau,\\
K_l, & z_t<\tau,
\end{cases}
\qquad K_s<K_l,
\end{equation}
where \(\tau\) is the validation-selected threshold, and \(K_s,K_l\) are the short and long budgets. Higher plan roughness receives \(K_s\); otherwise, \(K_l\) applies. The budget remains fixed until refresh and counts reuse updates only, giving a maximum refresh interval of \(K_t+1\) updates. A refresh at \(t\) is followed by reuse updates \(t+1,\ldots,t+K_t\), with the next refresh due by \(t+K_t+1\). Fig.~\ref{fig:bounded_refresh} illustrates the aggregation of action variation and the resulting budget assignment.

The output horizon and refresh interval are separate parameters. Each update integrates \(H\) commands from the current measured state to produce an \(H\Delta t\)-second trajectory. Recurrent correction and tail completion sustain these updates throughout reuse, including intervals longer than the original visual prediction horizon.

Each sequence starts with an empty buffer and a plan refresh. A subsequent refresh replaces the buffer when it is unavailable or its reuse budget is exhausted. Between refreshes, execution uses \(B_{t-1}\) and current measured motion without reading a new image. Parameters are fixed after validation. Candidate budgets are ranked by validation FDE, then ADE and refresh rate, subject to the constraints relative to fixed-interval execution.

\section{Experiments}
\label{sec:experiments}

The experiments test whether motion-conditioned plan maintenance can reduce visual inference while preserving trajectory accuracy in off-road driving. Comparisons across planning systems and frozen backbones establish overall performance, while controlled ablations distinguish the effects of state correction, visual initialization, and refresh scheduling. Cross-sequence tests and independent model requests then examine whether the accuracy and inference savings persist beyond the shared-prediction replay.

\subsection{Experimental Setup}

\textbf{Data and Protocol.}
Experiments use ORAD-3D~\cite{orad3d2025} and RELLIS-3D~\cite{rellis3d2020}, with 636 evaluation frames from 11 sequences and 650 frames from three sequences, respectively. Each frame anchors a motion history and its future ground-truth (GT) trajectory. ORAD-3D separates validation and test sequences; the RELLIS-3D temporal splits have disjoint evaluation timestamps and supporting history/future indices. Evaluation follows logged-state open-loop replay: measured motion is read at every update, independently of the preceding prediction, and the buffer resets at sequence boundaries. Within each backbone, all policies share the same evaluation frames and per-update processing. Every-update execution admits a fresh plan at each timestamp. Fixed-interval execution retains the recurrent interface with a constant refresh budget, while FastRoverVLA selects the budget from plan variation. Both reuse policies update the trajectory at every timestamp, including updates without a visual prediction. The matched comparisons use only motion history available at the current timestamp. Controlled replay reads the same cached predictions without invoking the model, and $U/R$ counts updates per admitted plan. Independent real-request runs generate their own plans on the same logged states.

\textbf{Evaluation Metrics.}
Each trajectory is evaluated at six ego-relative positions, spaced 0.5~s apart over 3~s. Average displacement error (ADE) averages the six position errors, and final displacement error (FDE) measures the 3~s endpoint error. Intermediate $L_2$ errors are reported at 1 and 2~s. Root mean squared error (RMSE) is computed from coordinate errors within each sample and averaged across frames; P95FDE is the 95th percentile of FDE. All errors are in meters. Sequence-level intervals use 10,000 paired whole-sequence bootstrap resamples, weighting sequence means equally to account for temporal dependence within recordings. ORAD-3D uses the preceding displacement as its reference heading. RELLIS-3D retains the archived centered-heading convention, with a past-only check in Section~\ref{sec:heading_check}.

\textbf{Implementation.}
All backbone comparisons use the same evaluation frames, GT trajectories, and execution operators. Model weights are frozen, and execution parameters are fixed after validation. The default Qwen3-VL configuration uses $\beta=0.65$ and $\delta_v=0.25$~m/s, with $(K_s,K_l,\tau,\alpha)=(12,32,0.1463,0.40)$ on ORAD-3D and $(8,24,1.2074,0.65)$ on RELLIS-3D; thresholds are rounded. Fixed reuse budgets are 12 and 8, respectively. Runtime measurements use one RTX~4090, BF16 weights, $448\times448$ images, and temperature zero. Request cost is cumulative model-request processing time per logged update under serial replay, excluding initialization, image preparation, and warm-up.

\begin{table}[!htbp]
\centering\ResultTableStyle
\caption{RELLIS-3D system comparison under the input and horizon adaptations in Section~\ref{sec:system_adaptations}. FastRoverVLA has the lowest displayed errors. Frozen weights; 3~s horizon; errors: m ($\downarrow$); bold: column minima.}
\label{tab:efficiency}
\begin{tabular}{@{}L{\ResultLabelWidth}*{6}{C{\ResultSystemMetricWidth}}@{}}
\toprule
\multirow{2}{*}{\textbf{Method}} & \multicolumn{3}{c}{\textbf{L\textsubscript{2} error}$\downarrow$} & \multirow{2}{*}{\textbf{ADE}$\downarrow$} & \multirow{2}{*}{\textbf{RMSE}$\downarrow$} & \multirow{2}{*}{\makecell{\textbf{P95FDE}\\$\downarrow$}} \\
\cmidrule(lr){2-4}
& \textbf{1 s} & \textbf{2 s} & \textbf{3 s} & & & \\
\midrule
OFF-EMMA~\cite{offemma2026} & 0.168 & 0.428 & 0.805 & 0.395 & 0.336 & 2.068 \\
ReCogDrive~\cite{recogdrive2026} & 0.292 & 0.693 & 1.365 & 0.692 & 0.586 & 3.306 \\
AutoVLA~\cite{autovla2026} & 0.293 & 1.217 & 3.089 & 1.235 & 1.159 & 5.293 \\
SimLingo~\cite{renz2025simlingo} & 2.411 & 8.201 & 14.067 & 6.955 & 5.951 & 20.018 \\
DriveMoE~\cite{drivemoe2026} & 1.347 & 3.763 & 6.219 & 3.206 & 2.684 & 10.480 \\
Drive-$\pi_0$~\cite{drivemoe2026} & 0.857 & 2.958 & 5.147 & 2.521 & 2.170 & 10.647 \\
\midrule
\textbf{FastRoverVLA} & \textbf{0.126} & \textbf{0.286} & \textbf{0.526} & \textbf{0.266} & \textbf{0.222} & \textbf{1.375} \\
\bottomrule
\end{tabular}
\end{table}

\textbf{Baseline Methods.}\label{sec:system_adaptations}
The comparison includes OFF-EMMA~\cite{offemma2026}, ReCogDrive~\cite{recogdrive2026}, AutoVLA~\cite{autovla2026}, SimLingo~\cite{renz2025simlingo}, DriveMoE, and Drive-$\pi_0$~\cite{drivemoe2026}. These independently developed driving systems provide reference points for evaluating FastRoverVLA as a complete planning method. Their inclusion extends the evaluation across different pretrained planning pipelines. Complementary experiments apply the proposed execution interface to four frozen backbones and compare execution policies using shared predictions. The system-level comparison therefore assesses overall trajectory accuracy, while the matched backbone experiments and component ablations identify the contribution of recurrent execution.

\subsection{Main Results}

\begin{figure}[!htbp]
\centering
\includegraphics[width=0.95\textwidth]{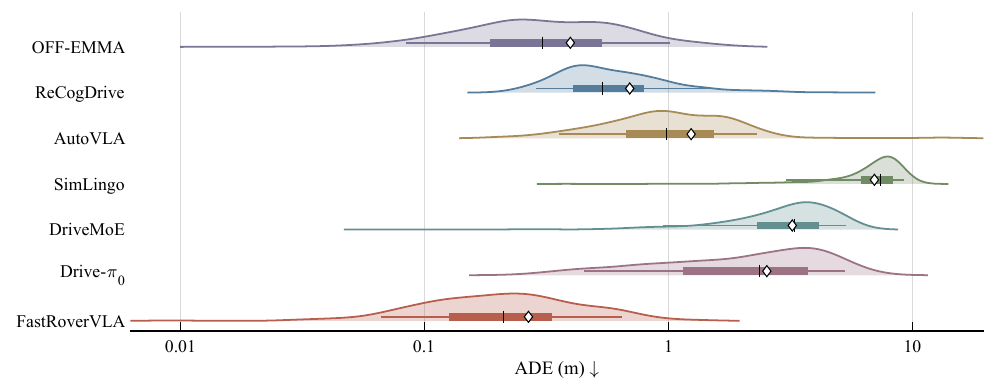}
\caption{Frame-level 3~s ADE on RELLIS-3D under the adaptations in Table~\ref{tab:efficiency}. Curves: peak-normalized densities in log-error space; thick bars: interquartile ranges; thin lines: 5th--95th percentiles; ticks: medians; diamonds: arithmetic means.}
\label{fig:model_comparison}
\end{figure}

FastRoverVLA has the lowest mean and upper-tail errors in Table~\ref{tab:efficiency}. Relative to OFF-EMMA, ADE decreases by 32.6\% and P95FDE by 33.5\%, showing comparable improvements in average and tail accuracy. At 0.5~s, within every model's native prediction horizon, mean position error is 0.070~m versus 0.087~m for OFF-EMMA. The short-horizon gain is therefore present before trajectory continuation affects the shorter outputs. Fig.~\ref{fig:model_comparison} also shows a lower central error range, although the distributions overlap across individual frames.

\begin{table}[!htbp]
\centering\ResultTableStyle
\caption{Recurrent execution improves all eight backbone--dataset pairs under shared frozen predictions. Errors: m ($\downarrow$); U/R: updates per admitted plan ($\uparrow$). Bold: within-backbone displayed optima, including ties.}
\label{tab:backbones}
\begin{tabular}{@{}L{\ResultBackboneWidth}L{\ResultExecutionWidth}*{8}{C{\ResultMetricWidth}}@{}}
\toprule
\multirow{2}{*}{\textbf{Backbone}} & \multirow{2}{*}{\textbf{Execution}} & \multicolumn{4}{c}{\textbf{ORAD-3D}} & \multicolumn{4}{c}{\textbf{RELLIS-3D}} \\
\cmidrule(lr){3-6}\cmidrule(lr){7-10}
& & \textbf{ADE}$\downarrow$ & \textbf{FDE}$\downarrow$ & \makecell{\textbf{P95FDE}\\$\downarrow$} & \textbf{U/R}$\uparrow$ & \textbf{ADE}$\downarrow$ & \textbf{FDE}$\downarrow$ & \makecell{\textbf{P95FDE}\\$\downarrow$} & \textbf{U/R}$\uparrow$ \\
\midrule
\multirow{3}{*}{Qwen3-VL-2B} & Every update & 0.384 & 0.726 & 1.736 & 1.0 & 0.387 & 0.820 & 1.933 & 1.0 \\
 & Fixed interval & 0.330 & 0.599 & 1.478 & 12.2 & 0.289 & 0.584 & 1.507 & 8.8 \\
 & \textbf{FastRoverVLA} & \textbf{0.327} & \textbf{0.593} & \textbf{1.445} & \textbf{17.2} & \textbf{0.266} & \textbf{0.526} & \textbf{1.375} & \textbf{21.0} \\
\midrule
\multirow{3}{*}{MiniCPM-V-2.6} & Every update & 0.390 & 0.740 & 1.732 & 1.0 & 0.457 & 0.905 & 1.886 & 1.0 \\
 & Fixed interval & 0.327 & 0.595 & 1.478 & 12.2 & 0.292 & 0.574 & 1.454 & 8.8 \\
 & \textbf{FastRoverVLA} & \textbf{0.324} & \textbf{0.589} & \textbf{1.455} & \textbf{18.2} & \textbf{0.282} & \textbf{0.555} & \textbf{1.394} & \textbf{21.0} \\
\midrule
\multirow{3}{*}{Qwen2.5-VL-3B} & Every update & 0.376 & 0.705 & 1.686 & 1.0 & 0.338 & 0.696 & 1.636 & 1.0 \\
 & Fixed interval & \textbf{0.332} & \textbf{0.601} & \textbf{1.481} & \textbf{12.2} & 0.278 & 0.556 & 1.480 & 8.8 \\
 & \textbf{FastRoverVLA} & \textbf{0.332} & \textbf{0.601} & \textbf{1.481} & \textbf{12.2} & \textbf{0.268} & \textbf{0.529} & \textbf{1.382} & \textbf{13.8} \\
\midrule
\multirow{3}{*}{InternVL2.5-2B} & Every update & 0.377 & 0.712 & 1.690 & 1.0 & 0.354 & 0.745 & 1.839 & 1.0 \\
 & Fixed interval & \textbf{0.331} & \textbf{0.601} & \textbf{1.464} & 12.2 & 0.287 & 0.582 & 1.531 & 8.8 \\
 & \textbf{FastRoverVLA} & 0.332 & \textbf{0.601} & \textbf{1.464} & \textbf{13.5} & \textbf{0.266} & \textbf{0.527} & \textbf{1.390} & \textbf{15.5} \\
\bottomrule
\end{tabular}
\end{table}

Recurrent execution lowers ADE and FDE across all eight backbone--dataset pairs in Table~\ref{tab:backbones}. For Qwen3-VL, FDE falls by 18.2\% on ORAD-3D and 35.8\% on RELLIS-3D, accompanied by P95FDE reductions of 16.8\% and 28.9\%. Fixed-interval reuse also improves accuracy, confirming that fewer predictions can remain effective when the retained actions are corrected. On RELLIS-3D, plan-conditioned refresh further reduces FDE by 3.3--9.9\% relative to fixed intervals and increases the number of updates served by each plan. Its additional accuracy gain is smaller on ORAD-3D, where Qwen2.5-VL has identical fixed and adaptive results.

\subsection{Ablation Study}

\begin{table}[!htbp]
\centering\ResultTableStyle
\caption{State alignment and write-back improve reuse. Upper block: shared refresh times; the no-reuse-correction control also shares refresh-time processing. Lower block: no requests or scheduled refresh. Errors: m ($\downarrow$); bold: displayed column minima, including ties.}
\label{tab:mechanism}
\begin{tabular}{@{}L{\ResultLabelWidth}*{8}{C{\ResultMetricWidth}}@{}}
\toprule
\multirow{2}{*}{\textbf{Control}} & \multicolumn{4}{c}{\textbf{ORAD-3D}} & \multicolumn{4}{c}{\textbf{RELLIS-3D}} \\
\cmidrule(lr){2-5}\cmidrule(lr){6-9}
& \textbf{ADE}$\downarrow$ & \textbf{FDE}$\downarrow$ & \textbf{RMSE}$\downarrow$ & \makecell{\textbf{P95FDE}\\$\downarrow$} & \textbf{ADE}$\downarrow$ & \textbf{FDE}$\downarrow$ & \textbf{RMSE}$\downarrow$ & \makecell{\textbf{P95FDE}\\$\downarrow$} \\
\midrule
\multicolumn{9}{@{}l}{\textbf{Shared VLA-derived refresh times}} \\
\textbf{FastRoverVLA} & 0.327 & 0.593 & 0.267 & 1.445 & 0.266 & 0.526 & 0.222 & 1.375 \\
Direct stale reuse & 0.500 & 0.906 & 0.404 & 2.260 & 0.471 & 0.901 & 0.387 & 2.646 \\
No correction during reuse & 0.507 & 0.924 & 0.411 & 2.330 & 0.455 & 0.876 & 0.375 & 2.511 \\
No per-update alignment ($\beta=1$) & 0.454 & 0.852 & 0.374 & 2.064 & 0.419 & 0.844 & 0.353 & 2.360 \\
No correction write-back & 0.396 & 0.728 & 0.323 & 1.776 & 0.371 & 0.738 & 0.310 & 1.907 \\
No fresh-plan blend ($\alpha=1$) & 0.327 & 0.593 & 0.267 & \textbf{1.440} & 0.266 & 0.527 & 0.223 & 1.380 \\
No speed bound & \textbf{0.326} & \textbf{0.592} & \textbf{0.266} & 1.459 & 0.267 & 0.527 & 0.223 & 1.375 \\
Shuffled plan & 0.341 & 0.609 & 0.277 & 1.548 & 0.277 & 0.548 & 0.231 & 1.381 \\
Kinematic plan & 0.329 & 0.597 & 0.269 & 1.450 & 0.270 & 0.537 & 0.226 & 1.375 \\
Null plan & 0.394 & 0.733 & 0.324 & 2.043 & 0.335 & 0.646 & 0.278 & 1.615 \\
\midrule
\multicolumn{9}{@{}l}{\textbf{Motion-only controls: no visual requests or refresh schedule}} \\
Constant current motion & 0.407 & 0.775 & 0.335 & 1.780 & 0.434 & 0.937 & 0.375 & 2.066 \\
Persistent motion-only & 0.330 & 0.599 & 0.270 & 1.477 & \textbf{0.261} & \textbf{0.515} & \textbf{0.218} & \textbf{1.367} \\
\bottomrule
\end{tabular}
\end{table}

Table~\ref{tab:mechanism} tests the execution components at matched refresh times. Removing alignment increases FDE by 43.7\% on ORAD-3D and 60.5\% on RELLIS-3D; removing write-back increases it by 22.8\% and 40.3\%. Current-state correction and its persistence therefore make distinct contributions. Fresh-plan blending and speed bounds have smaller effects. Matched plans outperform shuffled and null initializations, but kinematic initialization comes within 0.004/0.011~m FDE of the full method. Persistent motion-only execution is similarly accurate on ORAD-3D and slightly more accurate on RELLIS-3D, identifying recurrent motion correction as the dominant source of the local prediction gain.

\begin{figure}[!htbp]
\centering
\includegraphics[width=0.93\textwidth]{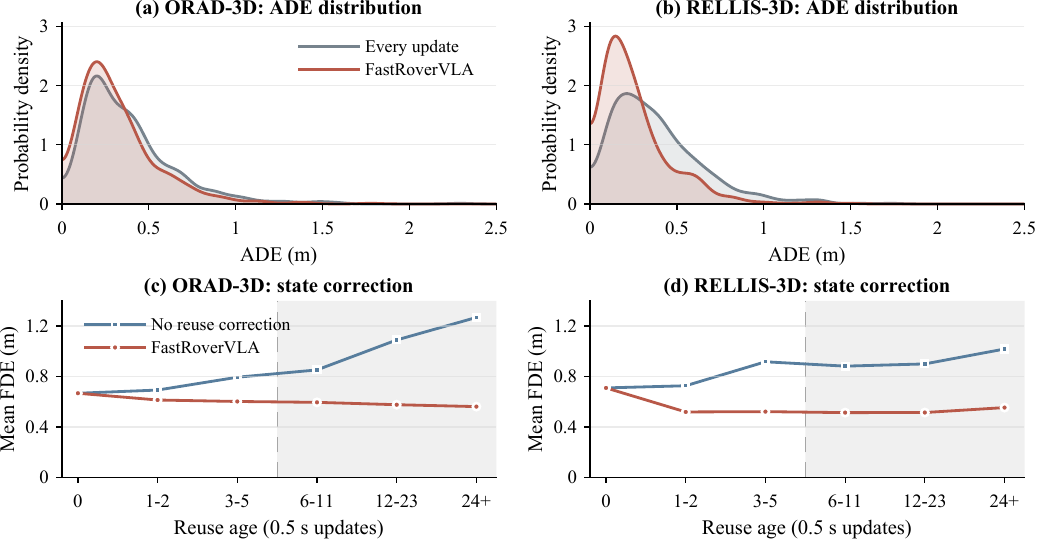}
\caption{Matched Qwen3-VL replay. (a,b) ADE densities use a shared within-dataset bandwidth and reflection at zero. (c,d) Mean FDE by reuse age; both policies have identical refresh-time trajectories, but the control omits subsequent correction. Shading marks ages beyond the six-action horizon.}
\label{fig:results_overview}
\end{figure}

Fig.~\ref{fig:results_overview}(c,d) compares policies with identical refresh-time plans. The gap at older reuse ages shows the benefit of continuing to correct the retained actions, including beyond the original six-action horizon. Each curve averages the frames reaching a given age.

The image control in Table~\ref{tab:image_content} holds motion history and the prompt fixed while using true, gray, or cross-sequence donor images. Each case starts from a fresh plan. Turning cases require speed of at least 0.5~m/s, absolute yaw rate of at most $3^\circ$/s, and at least $15^\circ$ between valid first and last GT trajectory segments; GT is used only for this grouping. True images reduce FDE by 6.9\% versus gray inputs and 4.3\% versus donor inputs on the 14 RELLIS-3D turning frames. The seven ORAD-3D turning frames and the full test sets do not show the same advantage. Visual initialization thus has a conditional contribution in this diagnostic.

\begin{table}[!htbp]
\centering
\caption{Image-content controls at fresh-plan initialization. Turning subsets: Section~\ref{sec:image_diagnostic}. Errors: m ($\downarrow$); bold: dataset/group minima.}
\label{tab:image_content}
\begingroup
\small
\setlength{\tabcolsep}{3pt}
\renewcommand{\arraystretch}{1.03}
\begin{tabular*}{\columnwidth}{@{\extracolsep{\fill}}lcccc@{}}
\toprule
\multirow{2}{*}{\textbf{Image input}} & \multicolumn{2}{c}{\textbf{All frames}} & \multicolumn{2}{c}{\textbf{Turning subset}}\\
\cmidrule(lr){2-3}\cmidrule(l){4-5}
 & \textbf{ADE} & \textbf{FDE} & \textbf{ADE} & \textbf{FDE}\\
\midrule
\multicolumn{5}{c}{\textbf{ORAD-3D}}\\
True & 0.385 & 0.727 & 0.468 & 1.014\\
Gray & \textbf{0.384} & \textbf{0.726} & \textbf{0.464} & \textbf{1.005}\\
Donor & 0.385 & 0.728 & 0.465 & 1.008\\
\midrule
\multicolumn{5}{c}{\textbf{RELLIS-3D}}\\
True & 0.391 & 0.831 & \textbf{0.515} & \textbf{1.080}\\
Gray & \textbf{0.386} & \textbf{0.820} & 0.539 & 1.159\\
Donor & 0.387 & 0.822 & 0.523 & 1.128\\
\bottomrule
\end{tabular*}
\endgroup
\end{table}

\label{sec:image_diagnostic}

\begin{figure}[!htbp]
\centering
\includegraphics[width=1.00\textwidth]{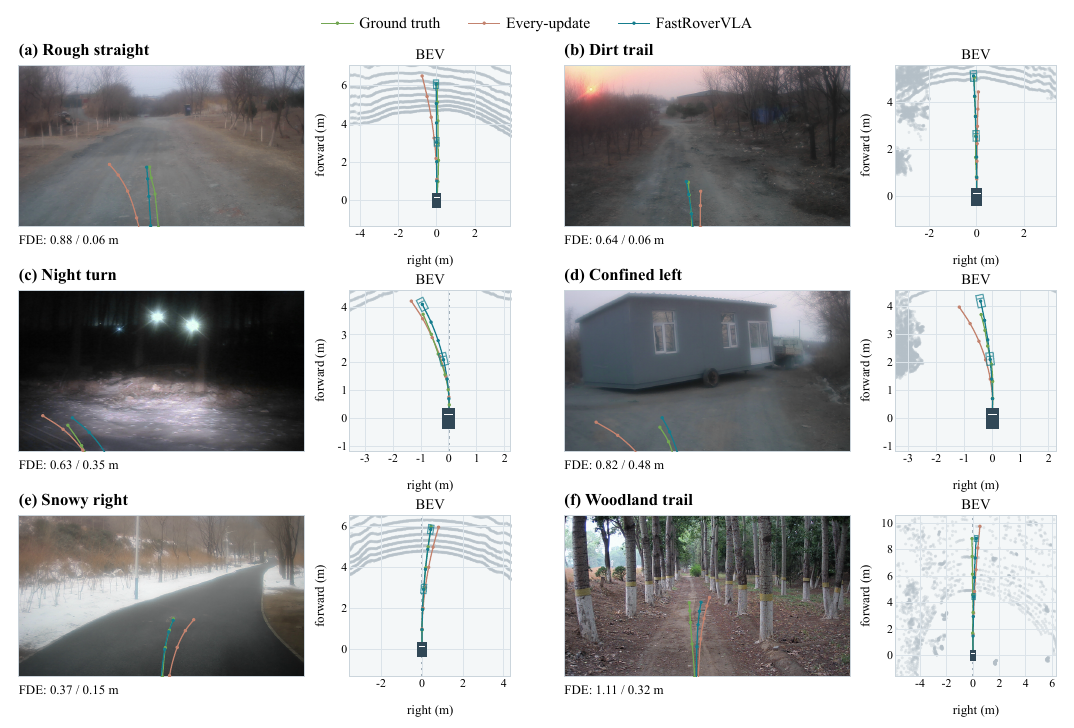}
\caption{ORAD-3D trajectories from matched Qwen3-VL predictions. Camera and BEV views show six waypoints; curves interpolate them for display. Boxes: schematic ego poses; gray points: synchronized LiDAR. Camera projections use per-frame calibration and a 0.65~m display lift. Labels give every-update / FastRoverVLA FDE.}
\label{fig:qualitative_execution}
\end{figure}

Fig.~\ref{fig:qualitative_execution} compares trajectories in camera and metric BEV views across straight travel and turns. FastRoverVLA follows the recorded path more closely on the rough straight and confined left turn, where every-update predictions accumulate lateral deviation. The BEV views distinguish path direction from endpoint distance, while the camera views retain the surrounding terrain and visibility conditions. Further examples can be explored at \url{https://adept-org.github.io/FastRoverVLA/}.

\subsection{Robustness and Sensitivity Analysis}
\label{sec:generalization}

\begin{table}[!htbp]
\centering\ResultTableStyle
\caption{Refresh timing, signals, and reuse caps in frozen-Qwen3 replay. R/C/Q: initial-action residual, output incompleteness, and plan roughness. Errors: m ($\downarrow$); rate: admissions/update (\%); bold: displayed block minima.}
\label{tab:refresh}
\begin{tabular}{@{}L{\ResultLabelWidth}*{8}{C{\ResultMetricWidth}}@{}}
\toprule
\multirow{2}{*}{\textbf{Setting}} & \multicolumn{4}{c}{\textbf{ORAD-3D}} & \multicolumn{4}{c}{\textbf{RELLIS-3D}} \\
\cmidrule(lr){2-5}\cmidrule(lr){6-9}
& \textbf{ADE}$\downarrow$ & \textbf{FDE}$\downarrow$ & \makecell{\textbf{P95FDE}\\$\downarrow$} & \textbf{Rate}(\%)$\downarrow$ & \textbf{ADE}$\downarrow$ & \textbf{FDE}$\downarrow$ & \makecell{\textbf{P95FDE}\\$\downarrow$} & \textbf{Rate}(\%)$\downarrow$ \\
\midrule
\multicolumn{9}{@{}l}{\textbf{Timing and cross-dataset rule transfer}} \\
Fixed interval & 0.330 & 0.599 & 1.478 & 8.2 & 0.289 & 0.584 & 1.507 & 11.4 \\
Uniform, matched budget & 0.329 & 0.597 & 1.519 & 5.8 & 0.268 & 0.530 & 1.377 & \textbf{4.8} \\
Local plan-conditioned rule & \textbf{0.327} & \textbf{0.593} & \textbf{1.445} & 5.8 & \textbf{0.266} & \textbf{0.526} & \textbf{1.375} & \textbf{4.8} \\
Transferred frozen rule & 0.331 & 0.603 & 1.531 & \textbf{5.3} & 0.272 & 0.541 & 1.391 & 5.7 \\
\midrule
\multicolumn{9}{@{}l}{\textbf{Validation-selected refresh signals}} \\
$Q$ only & \textbf{0.327} & \textbf{0.593} & \textbf{1.445} & 5.8 & \textbf{0.266} & \textbf{0.526} & \textbf{1.375} & \textbf{4.8} \\
$R+C$ & 0.334 & 0.607 & 1.544 & 6.1 & 0.271 & 0.539 & 1.394 & 5.2 \\
$R+Q$ & 0.329 & 0.597 & 1.508 & 7.7 & 0.270 & 0.537 & 1.394 & \textbf{4.8} \\
$C+Q$ & 0.331 & 0.599 & 1.477 & \textbf{4.9} & \textbf{0.266} & \textbf{0.526} & \textbf{1.375} & \textbf{4.8} \\
$R+C+Q$ & 0.334 & 0.606 & 1.544 & 6.0 & 0.270 & 0.537 & 1.394 & \textbf{4.8} \\
\midrule
\multicolumn{9}{@{}l}{\textbf{Maximum reuse budget (updates)}} \\
$K_{\max}=6$ & 0.342 & 0.623 & 1.548 & 15.1 & 0.290 & 0.584 & 1.502 & 14.3 \\
$K_{\max}=8$ & 0.335 & 0.610 & 1.467 & 11.8 & 0.289 & 0.584 & 1.507 & 11.4 \\
$K_{\max}=12$ & 0.330 & 0.599 & 1.478 & 8.2 & 0.272 & 0.543 & 1.397 & 8.2 \\
$K_{\max}=16$ & 0.328 & 0.595 & 1.508 & 7.5 & 0.276 & 0.552 & 1.398 & 6.8 \\
$K_{\max}=24$ & \textbf{0.327} & 0.596 & 1.473 & 6.9 & \textbf{0.266} & \textbf{0.526} & \textbf{1.375} & \textbf{4.8} \\
$K_{\max}=32$ & \textbf{0.327} & \textbf{0.593} & \textbf{1.445} & \textbf{5.8} & \textbf{0.266} & \textbf{0.526} & \textbf{1.375} & \textbf{4.8} \\
\bottomrule
\end{tabular}
\end{table}

Table~\ref{tab:refresh} examines the accuracy retained at different refresh budgets. On RELLIS-3D, the plan-conditioned rule reduces FDE by 9.9\% and admissions by 58.1\% relative to fixed intervals. The offline uniform schedule at the same admission count remains within 0.004~m FDE on both datasets, showing that these sparse budgets tolerate changes in refresh timing. The transferred rule also retains lower errors than every-update execution. Adding residual or incompleteness signals does not consistently improve ADE or FDE over plan roughness alone. Larger reuse caps reduce admissions, while upper-tail error fluctuates rather than improving monotonically.

\begin{table}[!htbp]
\centering\ResultTableStyle
\caption{Sequence-balanced error reductions (every-update minus FastRoverVLA, m $\uparrow$). Intervals: 10,000 whole-sequence bootstrap resamples; improved: sequences with lower error.}
\label{tab:statistics}
\begin{tabular*}{\textwidth}{@{\extracolsep{\fill}}llccc@{}}
\toprule
\textbf{Dataset} & \textbf{Metric} & \textbf{Reduction} & \textbf{Interval}(95\%) & \textbf{Improved} \\
\midrule
ORAD-3D & ADE & 0.055 & [0.030, 0.078] & 10/11 \\
 & FDE & 0.125 & [0.073, 0.175] & 10/11 \\
\midrule
RELLIS-3D & ADE & 0.111 & [0.092, 0.138] & 3/3 \\
 & FDE & 0.271 & [0.239, 0.328] & 3/3 \\
\bottomrule
\end{tabular*}
\end{table}

\begin{table}[!htbp]
\centering\ResultTableStyle
\caption{Past-only heading preserves the execution effect on RELLIS-3D. Only the scoring frame changes. Errors: m ($\downarrow$); bold: upper-panel column minima and lower-panel within-backbone minima.}
\label{tab:heading_check}
\begin{tabular*}{\textwidth}{@{\extracolsep{\fill}}lrrrr@{}}
\toprule
\multirow{2}{*}{\textbf{Controlled execution}} & \multicolumn{2}{c}{\textbf{Archived heading}} & \multicolumn{2}{c}{\textbf{Past-only heading}} \\
\cmidrule(lr){2-3}\cmidrule(lr){4-5}
& \textbf{ADE}$\downarrow$ & \textbf{FDE}$\downarrow$ & \textbf{ADE}$\downarrow$ & \textbf{FDE}$\downarrow$ \\
\midrule
Every update & 0.387 & 0.820 & 0.438 & 0.903 \\
Fixed interval & 0.289 & 0.584 & 0.339 & 0.666 \\
\textbf{FastRoverVLA} & 0.266 & 0.526 & 0.316 & 0.606 \\
No per-update alignment & 0.419 & 0.844 & 0.449 & 0.883 \\
No correction write-back & 0.371 & 0.738 & 0.414 & 0.805 \\
Kinematic initialization, matched refresh & 0.270 & 0.537 & 0.320 & 0.616 \\
Constant current motion & 0.434 & 0.937 & 0.483 & 1.018 \\
Persistent motion only, no refresh & \textbf{0.261} & \textbf{0.515} & \textbf{0.311} & \textbf{0.595} \\
\midrule
\textbf{Independent real-request trajectories} & \multicolumn{2}{c}{\textbf{Past-only ADE}} & \multicolumn{2}{c}{\textbf{Past-only FDE}} \\
\cmidrule(lr){2-3}\cmidrule(lr){4-5}
\textbf{Backbone} & \textbf{Every update} & \textbf{FastRoverVLA} & \textbf{Every update} & \textbf{FastRoverVLA} \\
\midrule
Qwen3-VL-2B & 0.441 & \textbf{0.315} & 0.914 & \textbf{0.605} \\
MiniCPM-V-2.6 & 0.412 & \textbf{0.318} & 0.834 & \textbf{0.612} \\
Qwen2.5-VL-3B & 0.389 & \textbf{0.318} & 0.781 & \textbf{0.611} \\
InternVL2.5-2B & 0.400 & \textbf{0.316} & 0.819 & \textbf{0.607} \\
\bottomrule
\end{tabular*}
\end{table}

\begin{table}[!htbp]
\centering\ResultTableStyle
\caption{Independent real-request execution improves accuracy and reduces cost relative to every-update prediction. Errors: m ($\downarrow$); Req.: requests/update (\%); C: request-processing ms/update. Bold: within-backbone minima, including ties.}
\label{tab:live}
\begin{tabular}{@{}L{\ResultBackboneWidth}L{\ResultExecutionWidth}*{8}{C{\ResultMetricWidth}}@{}}
\toprule
\multirow{2}{*}{\textbf{Backbone}} & \multirow{2}{*}{\textbf{Execution}} & \multicolumn{4}{c}{\textbf{ORAD-3D}} & \multicolumn{4}{c}{\textbf{RELLIS-3D}} \\
\cmidrule(lr){3-6}\cmidrule(lr){7-10}
& & \textbf{ADE}$\downarrow$ & \textbf{FDE}$\downarrow$ & \textbf{Req.}(\%)$\downarrow$ & \textbf{C}$\downarrow$ & \textbf{ADE}$\downarrow$ & \textbf{FDE}$\downarrow$ & \textbf{Req.}(\%)$\downarrow$ & \textbf{C}$\downarrow$ \\
\midrule
\multirow{3}{*}{Qwen3-VL-2B} & Every update & 0.385 & 0.727 & 100.0 & 355.4 & 0.391 & 0.831 & 100.0 & 443.9 \\
 & Fixed interval & 0.330 & 0.599 & 8.2 & 28.0 & 0.287 & 0.581 & 11.4 & 50.1 \\
 & \textbf{FastRoverVLA} & \textbf{0.328} & \textbf{0.596} & \textbf{6.0} & \textbf{21.4} & \textbf{0.266} & \textbf{0.526} & \textbf{4.8} & \textbf{21.3} \\
\midrule
\multirow{3}{*}{MiniCPM-V-2.6} & Every update & 0.397 & 0.754 & 100.0 & 1336.1 & 0.364 & 0.753 & 100.0 & 1392.8 \\
 & Fixed interval & 0.332 & 0.604 & 8.2 & 108.1 & 0.280 & 0.558 & 11.4 & 157.3 \\
 & \textbf{FastRoverVLA} & \textbf{0.329} & \textbf{0.597} & \textbf{6.0} & \textbf{79.8} & \textbf{0.269} & \textbf{0.533} & \textbf{4.8} & \textbf{66.2} \\
\midrule
\multirow{3}{*}{Qwen2.5-VL-3B} & Every update & 0.377 & 0.706 & 100.0 & 3461.2 & 0.339 & 0.697 & 100.0 & 3212.2 \\
 & Fixed interval & \textbf{0.332} & \textbf{0.601} & \textbf{8.2} & \textbf{246.6} & 0.277 & 0.555 & 11.4 & 356.2 \\
 & \textbf{FastRoverVLA} & \textbf{0.332} & \textbf{0.601} & \textbf{8.2} & 255.6 & \textbf{0.268} & \textbf{0.529} & \textbf{7.2} & \textbf{225.6} \\
\midrule
\multirow{3}{*}{InternVL2.5-2B} & Every update & 0.377 & 0.709 & 100.0 & 2102.2 & 0.353 & 0.742 & 100.0 & 2089.0 \\
 & Fixed interval & 0.331 & 0.601 & 8.2 & 185.7 & 0.284 & 0.574 & 11.4 & 236.0 \\
 & \textbf{FastRoverVLA} & \textbf{0.330} & \textbf{0.599} & \textbf{7.5} & \textbf{171.7} & \textbf{0.266} & \textbf{0.527} & \textbf{6.5} & \textbf{128.5} \\
\bottomrule
\end{tabular}
\end{table}

Error reduction persists when sequences receive equal weight (Table~\ref{tab:statistics}). The 95\% intervals exclude zero for both metrics, with lower FDE on 10 of 11 ORAD-3D sequences and all three RELLIS-3D sequences. The latter intervals describe variation among only three recordings. On an additional 931 ORAD-3D frames from disjoint sequences, the frozen rule also improves every-update execution, reaching 0.820~m FDE at a 6.6\% admission rate. The execution benefit therefore extends beyond the sequences used for calibration and the main test.


\label{sec:heading_check}
The execution gain persists when the RELLIS-3D reference heading uses past positions only (Table~\ref{tab:heading_check}). Rescoring the same trajectories gives a 32.9\% FDE reduction in controlled Qwen3-VL replay and reductions of 21.8--33.8\% across the four real-request backbones. Inputs, actions, and refresh schedules are unchanged. Removing alignment or write-back still increases error, so the component effects are preserved as well. The improvement consequently does not depend on the archived centered-heading convention.

\subsection{Inference Efficiency}
\label{sec:actual_requests}


FastRoverVLA reduces model requests by 91.8--95.2\% across the eight backbone--dataset pairs while also lowering ADE and FDE relative to every-update execution (Table~\ref{tab:live}). Qwen3-VL requests a fresh plan on 6.0\% of ORAD-3D updates and 4.8\% of RELLIS-3D updates, reducing cumulative request-processing time by factors of 16.6 and 20.8. Its 3~s FDE simultaneously decreases by 18.0\% and 36.7\%. The saving comes from avoiding repeated visual inference while motion-conditioned trajectory updates continue at every logged state.

\FloatBarrier
\section{Conclusion}

FastRoverVLA maintains decoded driving plans through recurrent motion correction, allowing visual planning and trajectory updates to run at different frequencies. State alignment, action consumption, and correction write-back preserve a usable speed--curvature buffer between bounded refreshes. Across four frozen backbones and two off-road datasets, the interface reduces trajectory error and model-request demand without retraining. Independent Qwen3-VL runs reduce requests by 94.0\% on ORAD-3D and 95.2\% on RELLIS-3D, with 3~s FDE reductions of 18.0\% and 36.7\% and request-processing savings of 16.6 and 20.8 times. The ablations show that accuracy depends on carrying motion corrections into later updates, while bounded refresh governs inference demand. The corrected action remainder enables later trajectory updates to use motion feedback accumulated since the last visual prediction. Closed-loop evaluation remains necessary to assess performance under state-estimation error and changing scene conditions.

\section*{Acknowledgments}
This work was supported by the National Natural Science
Foundation of China under Grant No. 52502496, the Beijing Natural
Science Foundation under Grant No. L2609087, and the Natural
Science Foundation of Chongqing, China under Grant No.
CSTB2025NSCQ-GPX0413.

{\small
\bibliographystyle{unsrt}
\bibliography{IEEEabrv,IROS,reference_updates}
}

\par\smallskip
\appendix
\section{Author Biographies}
\label{app:biographies}
\begingroup\setlength{\parskip}{0pt}

\noindent\begin{minipage}{\linewidth}\small
\begin{minipage}[t]{0.10\linewidth}\vspace{0pt}\centering
\includegraphics[width=0.57in,height=0.60in,keepaspectratio]{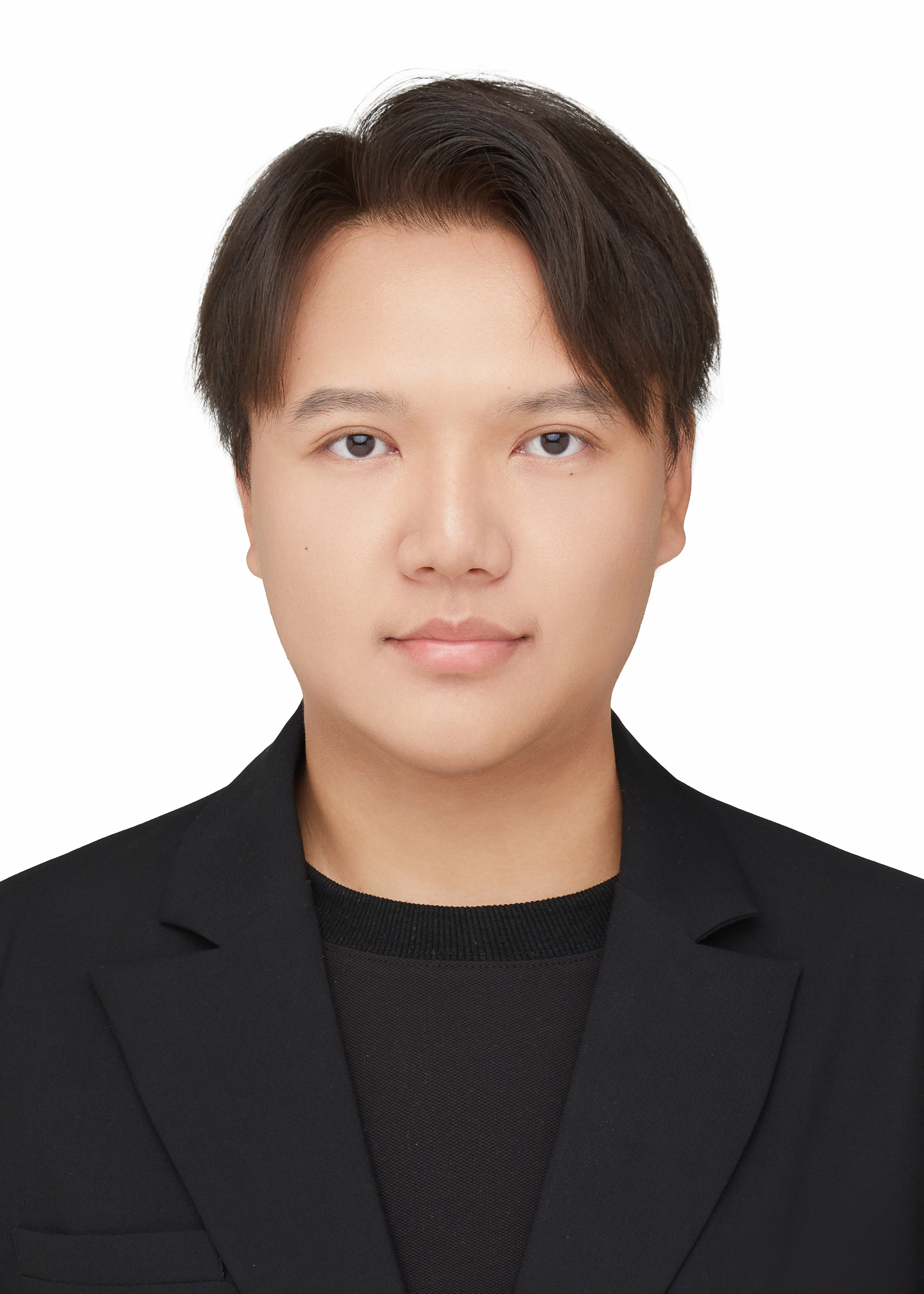}
\end{minipage}\hfill
\begin{minipage}[t]{0.88\linewidth}\vspace{0pt}
\textbf{Changtao Zhao} is currently pursuing the B.E. degree in mechatronic engineering at China Agricultural University, Beijing, China. He was a research intern with the School of Mechanical Engineering, Beijing Institute of Technology, Beijing, China. His research interests include autonomous driving, embodied intelligence, computer vision, and robotics.
\end{minipage}
\end{minipage}\par\smallskip

\noindent\begin{minipage}{\linewidth}\small
\begin{minipage}[t]{0.10\linewidth}\vspace{0pt}\centering
\includegraphics[width=0.57in,height=0.60in,keepaspectratio]{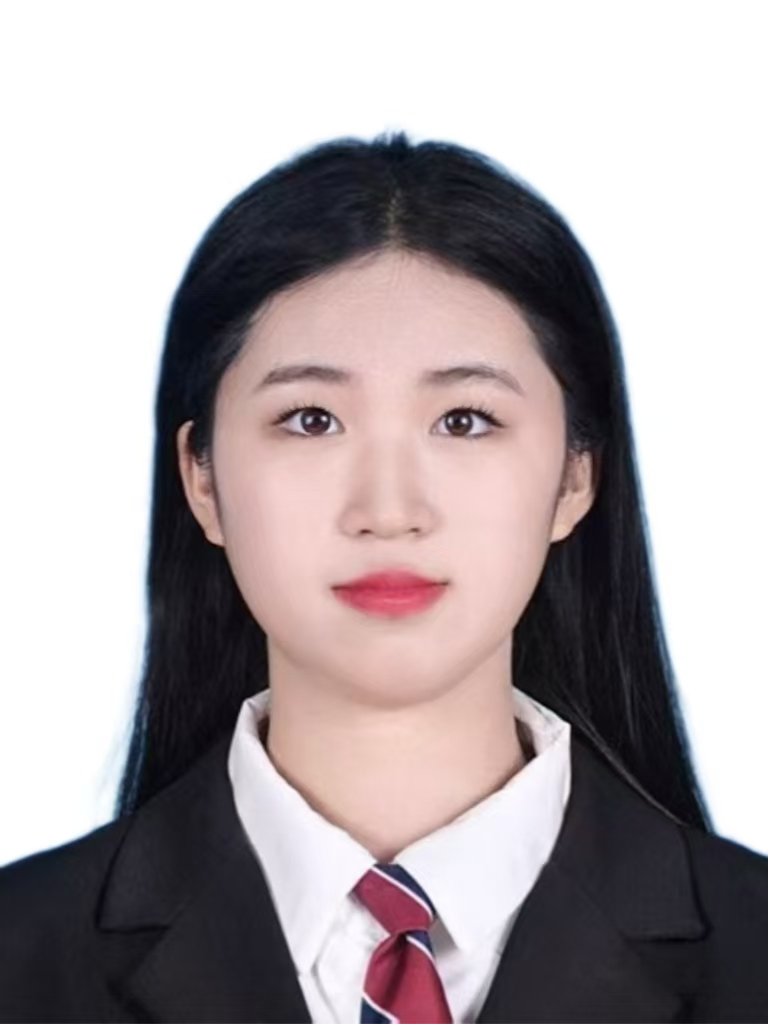}
\end{minipage}\hfill
\begin{minipage}[t]{0.88\linewidth}\vspace{0pt}
\textbf{Yingxue Guan} is currently pursuing the B.E. degree at China Agricultural University, Beijing, China. She was a research intern with the School of Mechanical Engineering, Beijing Institute of Technology, Beijing, China. Her research interests include computer vision, autonomous driving, mechanical design, and robotics.
\end{minipage}
\end{minipage}\par\smallskip

\noindent\begin{minipage}{\linewidth}\small
\begin{minipage}[t]{0.10\linewidth}\vspace{0pt}\centering
\includegraphics[width=0.57in,height=0.60in,keepaspectratio]{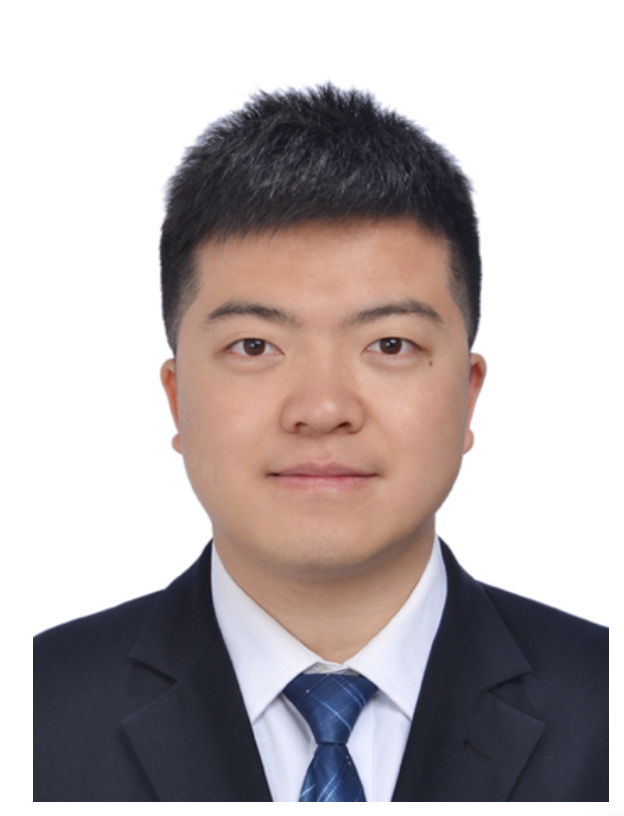}
\end{minipage}\hfill
\begin{minipage}[t]{0.88\linewidth}\vspace{0pt}
\textbf{Li Wang} received the Ph.D. degree in mechatronic engineering from Harbin Institute of Technology, China, in 2020. He was a postdoctoral researcher with the Department of Automotive Engineering, Tsinghua University, China. He is currently a Special Associate Researcher with the School of Mechanical Engineering, Beijing Institute of Technology, Beijing, China. His research interests include off-road autonomous driving, embodied intelligence, human--machine collaboration for intelligent unmanned systems, and 3-D scene reconstruction.
\end{minipage}
\end{minipage}\par\smallskip

\noindent\begin{minipage}{\linewidth}\small
\begin{minipage}[t]{0.10\linewidth}\vspace{0pt}\centering
\includegraphics[width=0.57in,height=0.60in,keepaspectratio]{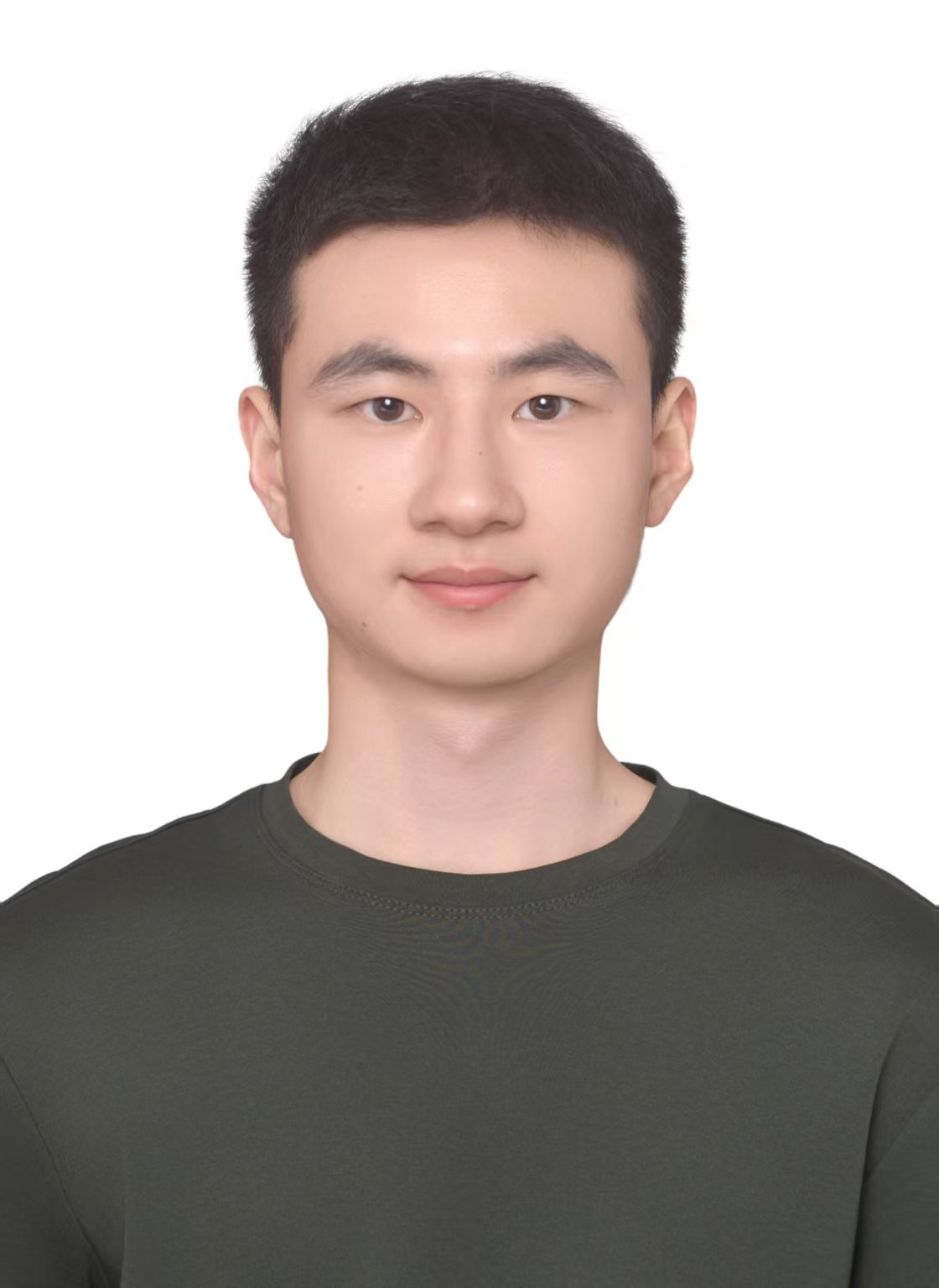}
\end{minipage}\hfill
\begin{minipage}[t]{0.88\linewidth}\vspace{0pt}
\textbf{Shijie Zhao} received the B.S. degree in mechanical engineering from Chang'an University, China, in 2024. He is currently pursuing the M.S. degree in mechanical engineering at Chang'an University, China. His research interests include autonomous driving and vision-language-action models.
\end{minipage}
\end{minipage}\par\smallskip

\noindent\begin{minipage}{\linewidth}\small
\begin{minipage}[t]{0.10\linewidth}\vspace{0pt}\centering
\includegraphics[width=0.57in,height=0.60in,keepaspectratio]{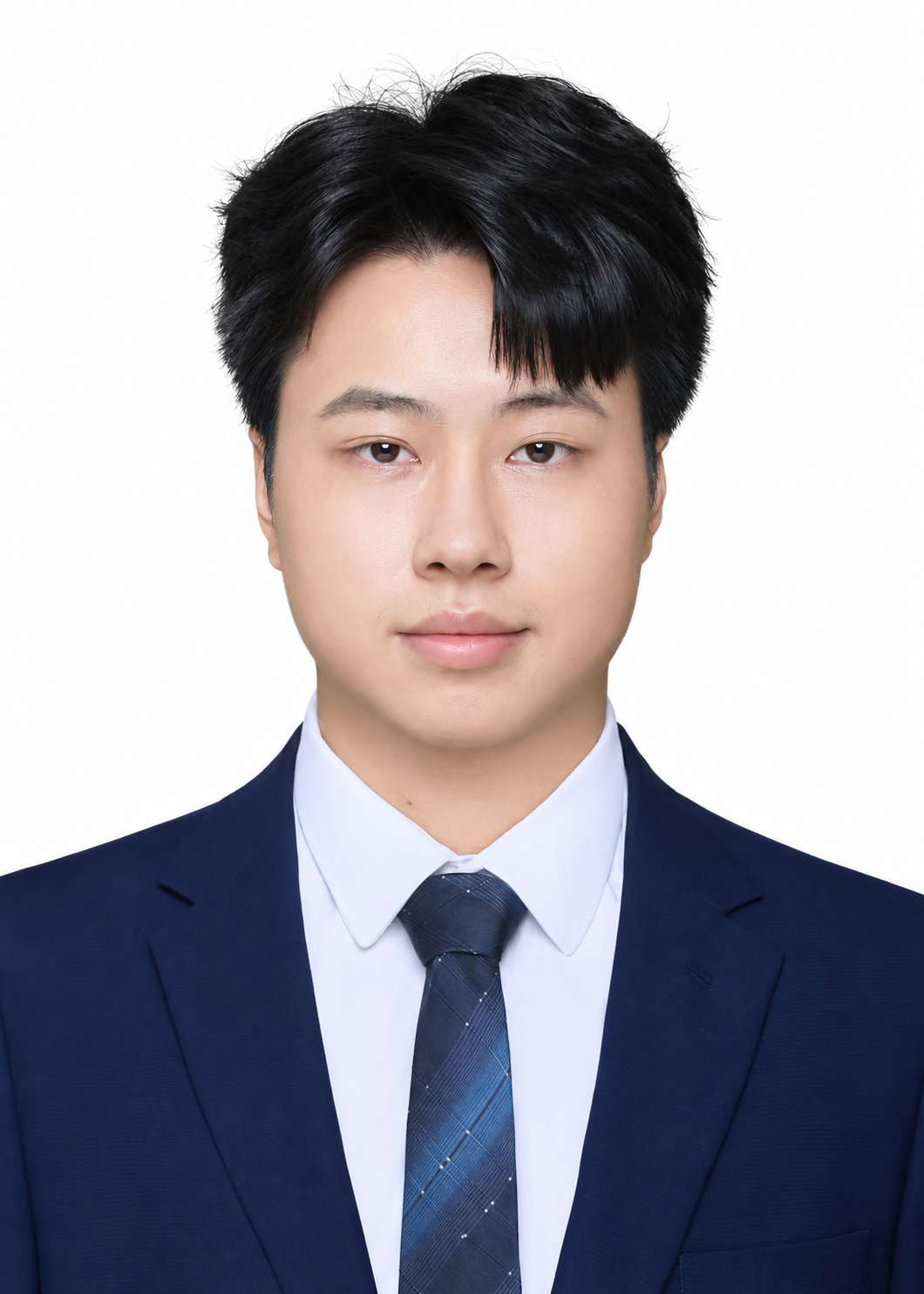}
\end{minipage}\hfill
\begin{minipage}[t]{0.88\linewidth}\vspace{0pt}
\textbf{Yuhang Zhang} studied for the B.S. degree in mechanical engineering at Chang'an University, China, from 2020 to 2024. Since 2024, he has been pursuing the M.S. degree in mechanical engineering at Chang'an University, China. His research interests include autonomous driving and world models.
\end{minipage}
\end{minipage}\par\smallskip

\noindent\begin{minipage}{\linewidth}\small
\begin{minipage}[t]{0.10\linewidth}\vspace{0pt}\centering
\includegraphics[width=0.57in,height=0.60in,keepaspectratio]{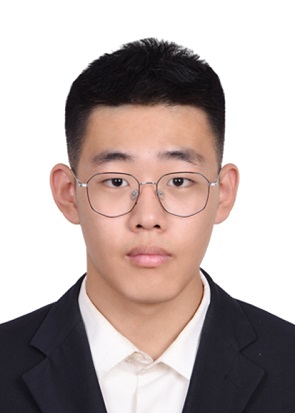}
\end{minipage}\hfill
\begin{minipage}[t]{0.88\linewidth}\vspace{0pt}
\textbf{Shengkai Zhang} received the B.S. degree in computer science and technology from Hebei University, China, in 2026. He is currently pursuing the M.S. degree in computer technology at Beijing Jiaotong University, China. His research interests include end-to-end autonomous driving, world models, and embodied intelligence.
\end{minipage}
\end{minipage}\par\smallskip

\noindent\begin{minipage}{\linewidth}\small
\begin{minipage}[t]{0.10\linewidth}\vspace{0pt}\centering
\includegraphics[width=0.57in,height=0.60in,keepaspectratio]{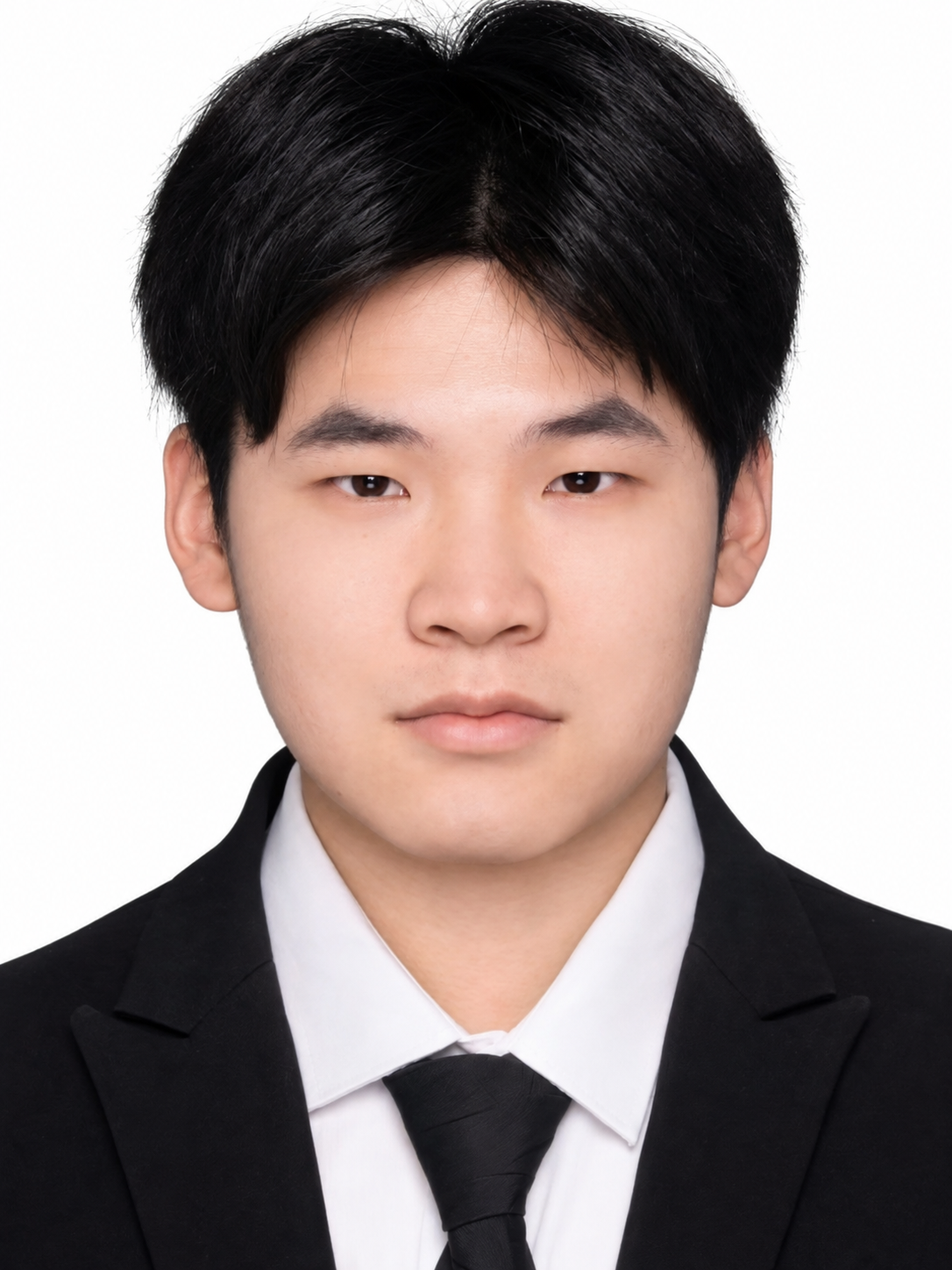}
\end{minipage}\hfill
\begin{minipage}[t]{0.88\linewidth}\vspace{0pt}
\textbf{Qing Wang} received the B.S. degree from Henan University, China, in 2026. He is currently pursuing the M.S. degree with the School of Artificial Intelligence, Yanshan University, China. His research interests include embodied agents, world models, and vision-language-action models.
\end{minipage}
\end{minipage}\par\smallskip

\noindent\begin{minipage}{\linewidth}\small
\begin{minipage}[t]{0.10\linewidth}\vspace{0pt}\centering
\includegraphics[width=0.57in,height=0.60in,keepaspectratio]{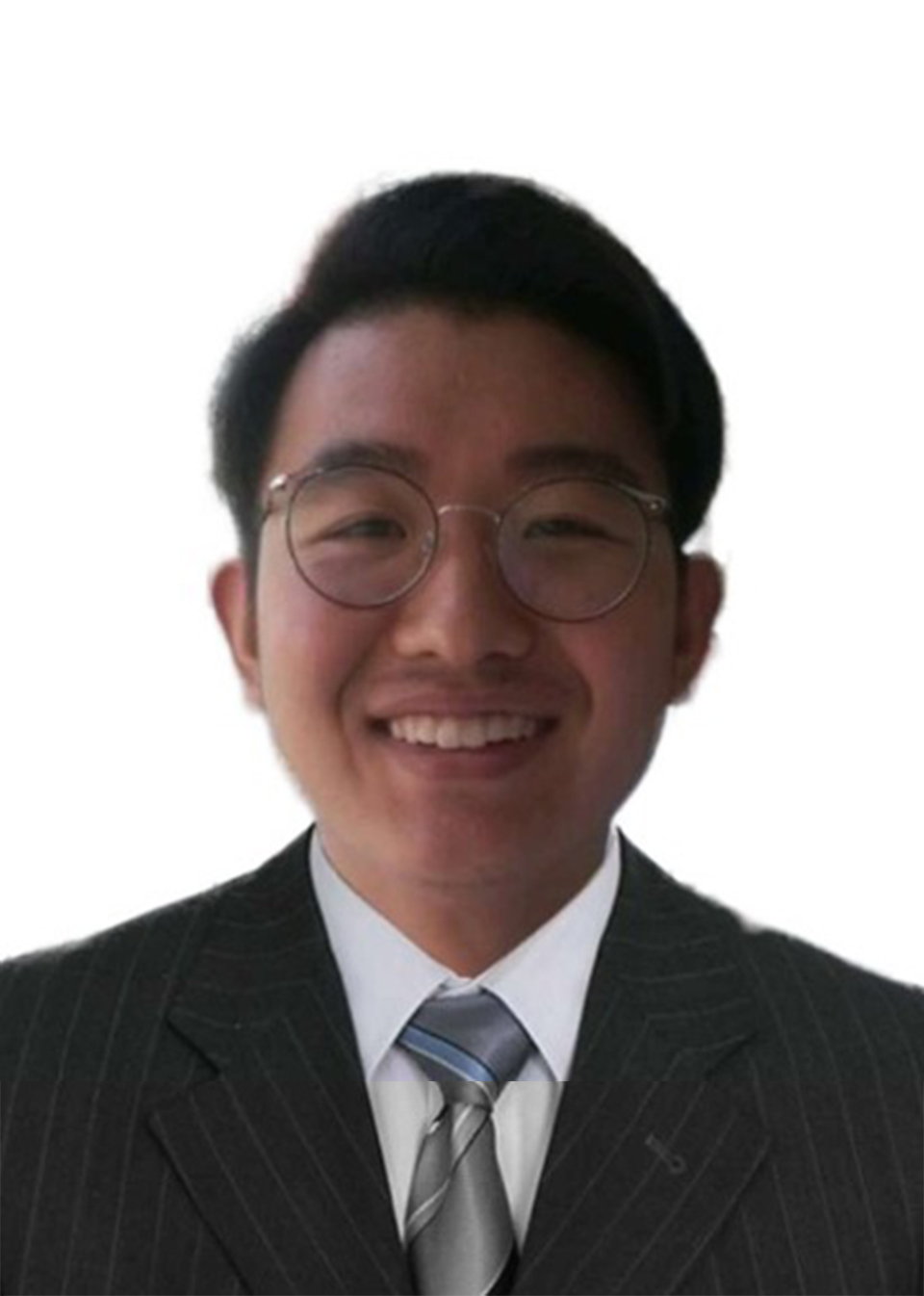}
\end{minipage}\hfill
\begin{minipage}[t]{0.88\linewidth}\vspace{0pt}
\textbf{Ziying Song} received the B.S. degree from Hebei Normal University of Science and Technology, China, in 2019, the M.S. degree from Hebei University of Science and Technology, China, in 2022, and the Ph.D. degree in computer science and technology from Beijing Jiaotong University, China, in March 2026. He is currently an Assistant Professor with the School of Artificial Intelligence, Yanshan University, China. His research interests include autonomous driving, robust perception, end-to-end autonomous driving, world models, embodied intelligence, and vision-language-action models.
\end{minipage}
\end{minipage}\par\smallskip

\endgroup
\end{document}